%% file: Rapid_SAT.tex
\documentclass{llncs}
\usepackage{amsmath} 
\usepackage{amssymb} 
\usepackage{graphicx}
\usepackage{algorithm}
\usepackage{algorithmic}
\usepackage{subfigure}
\usepackage{figsize}
\usepackage{verbatim} 
\usepackage{multirow}
\usepackage{array}
\usepackage{times}
\usepackage{multicol}
\usepackage[usenames]{color}
\usepackage{pstricks}
\usepackage{pst-node}

\begin{document}

\title{Towards Ultra Rapid Restarts\thanks{The second author is supported by the Dutch 
Organization for Scientific Research (NWO) under grant 617.023.611}}
\author{Shai Haim\inst{1} \and Marijn Heule\inst{2}}

\institute{University of New South Wales and NICTA, Sydney, Australia
\and
Delft University of Technology, Delft, The Netherlands}

\maketitle

\begin{abstract}
We observe a trend regarding restart strategies used in SAT solvers.
A few years ago, most state-of-the-art solvers restarted on average
after a few thousands of backtracks. Currently, restarting after a 
dozen backtracks results in much better performance. The main
reason for this trend is that heuristics and data structures have 
become more restart-friendly. We expect further 
continuation of this trend, so future SAT solvers will restart even 
more rapidly. Additionally, we present experimental results to support 
our observations.
\end{abstract}

\section{Introduction}

Restarts have been proposed for satisfiability (SAT) solvers to counter heavy-tail behavior~\cite{Gomes1998}.
Initially, branching heuristics were randomized to make sure that the search-tree would be different after 
each restart. Also, restarts should not be applied too frequently to guarantee that a solver can explore the entire
search-tree between two restarts in case a problem has no solutions. For modern conflict-driven clause learning
(CDCL) solvers~\cite{MSLM09HBSAT} this is no longer required. Decision heuristics are dynamic and updated after every 
conflict~\cite{MoskeWicz2001}. By recording conflict clauses, CDCL solvers can proof unsatisfiability even in case of ultra 
rapid restarts.

Nowadays, restarts have become an essential feature of CDCL solvers. Many different strategies have been 
studied and used~\cite{Luby1993,Walsh1999,Biere2008,Biere2008a,Biere2009b,Audemard2009,Pipatsrisawat2009,Ryvchin2008}. 
State-of-the-art SAT solvers tend to restart more and more frequently.
An explanation for this trend is that the heuristics and data-structures have become restart-friendly.
Therefore we decided to experiment with strategies that restart radically faster than commonly 
used in the CDCL solvers. The results show that these strategies are effective on the industrial
benchmarks of the SAT 2009 competition.

\section{Restarts}

Restart strategies have been used in SAT solvers for over a decade. First, we will provide an overview of their use in 
state-of-the-art solvers. Second, we will discuss two aspects of CDCL solvers, heuristics and data structures,
that influenced these strategies. Recent developments in these areas facilitate frequent restarts. Third, we argue 
that -- due to rapid restarts -- CDCL solvers have become complete local search solvers.

\subsection{A history of restart strategies}

Although currently all competitive CDCL solvers use restarts, this was not always the case. 
The solver {\tt grasp} \cite{Marques-Silva1996}, which first introduced clause learning in the 
context of satisfiability testing, did not use restarts in its original version. Following the work 
of Gomes et al. \cite{Gomes1998}, which demonstrated the effectiveness of restart for addressing 
issues arising from the heavy-tailed distribution, developers started equipping their solvers with 
fixed-size restart strategies. The solvers {\tt zChaff} \cite{MoskeWicz2001},  {\tt BerkMin} \cite{Goldberg2002} 
use rather frequent fixed restarts with restart sizes of 700 and 550 respectively, while the solver {\tt Siege} 
\cite{Ryan2004} uses a larger fixed restart size of 16,000 conflicts.

{\tt MiniSAT} 1.13 \cite{E'en2003} was the first to demonstrate the effectiveness of the geometric restart strategy suggested by Walsh \cite{Walsh1999}. Starting with a small first restart, the size of consecutive restarts grows geometrically.
A commonly used restart strategy in the recent years is based on a sequence of restart sizes suggested by Luby et al. \cite{Luby1993}. In their work the authors show that the suggested sequence is log optimal when the runtime distribution of the problems is unknown. In this strategy the length of restart $i$ is $u \cdot t_{i}$ when $u$ is a constant unit run and 
\[
t_{i} = 
\begin{cases} 
2^{k-1}\text{, } & \text{if } i=2^{k}-1 \\
t_{i-2^{k-1}+1}\text{, } & \text{if } 2^{k-1}\leq i < 2^{k}-1.
\end{cases}
\]
\\
Since unit runs are commonly short, solvers using the Luby restart strategy exhibit frequent restarts. The solvers {\tt Rsat} 2.0 \cite{Pipatsrisawat2007} and {\tt TiniSat} \cite{Huang2007a} use a unit run of 512 conflicts, while {\tt MiniSAT} 2.1 \cite{minisat21} and {\tt precoSAT} \cite{Biere2009b} use a shorter unit run of 100 conflicts. The solver {\tt picoSAT} \cite{Biere2008a} introduced a frequent restart strategy in which the restart length grows geometrically until it reaches a bound. At this point the restart sequence starts again and the bound grows geometrically. 
Another approach, which receives much attention lately, combines an underlying uniform restart strategy with a dynamic element which can induce, or suppress, restarts. The dynamic decision can be made according to variable agility \cite{Biere2008,Biere2008a,Biere2009b}, variety of decision levels in learnt clauses and backtrack sizes \cite{Audemard2009,Pipatsrisawat2009}, or using local search techniques \cite{Ryvchin2008}.

\subsection{Direction heuristics}

Direction heuristics select the value for decision variables. In theory, these heuristics can be very 
powerful: Perfect direction heuristics would result in a solver that never needs to backtrack to find 
a solution. If such a heuristic exists, which can be computed
in polynomial time, then $\mathcal{P} = \mathcal{NP}$. CDCL solvers use a variety of direction heursics.

For instance, {\tt zChaff} maintains 
two counters for each variable, one for true and one for false. These counters refer to the activity 
in recent conflicts. The sign of the highest counter is preferred. 
The direction heuristics in {\tt MiniSAT} are very minimalistic: It uses {\em negative branching}: i.e. 
the decision variable is always assigned to false. Although it might seem a bit arbitrary, it is not. 
Two properties of this heuristic contribute the fast performance. First, it consequently chooses the 
same sign. Therefore it keeps searching in the same search space.  Second, always branching on false 
is much better than always branching on true. The latter is an artifact of the encoding of most 
benchmark instances.
A direction heuristics technique called {\em phase-saving} was introduced in {\tt Rsat}.
Phase-saving assigns each decision variable to the value last forced by Boolean constraint 
propagation (BCP). In essence, this technique was already used in local search solvers. 
We will further discuss this in Section~\ref{sec:cls}.

The changes in direction heuristics can hardly be separated from the trend we ob- served for restart 
strategies. Rapid restarts only make sense in case the solver will not end up in a completely different 
search space again and again.  With the direction heuristics used in {\tt zChaff} this could easily happen. 
If for a high ranked variable both counters are almost equal than that variable can be flipped frequently.
While using negative branching it will happen less often. Yet as soon as a decision variable is chosen which BCP
mostly assigns to true, the search space becomes different. However, phase-saving is ideal for rapid
restarts since this direction heuristic ensures one hardly moves after a restart.

\subsection{Boolean constraint propagation}

Most of the computational cost of CDCL solvers is spent on BCP. Moskewicz et al.~\cite{MoskeWicz2001} state that in
most cases it is greater than 90\% of the total cost. This observation has consequences for rapid restart strategies:
If a solver would restart very frequently, say after every couple of conflicts, then it often has to go down the search-tree
all the way from the root. As a result, much more time will be spent on BCP slowing down the solver.

An important breakthrough in speeding up BCP is the introduction of the {\em watch 
literal data structure} in {\tt zChaff} \cite{MoskeWicz2001}. This data structure is now used in all state-of-the-art
CDCL solvers. It has been implemented very efficiently in {\tt MiniSAT}~\cite{E'en2003}. Recent 
improvements of this data structure were used in {\tt picoSAT}~\cite{Biere2008a}. Additionally, the relative burden
of BCP can be reduced by spending more time on reasoning techniques. For example by making conflict analysis
stronger. Two recent improvements in this direction are conflict clause minimization~\cite{minimazation} and conflict clause
(self-) subsumption~\cite{fly}.

Both developments influence the optimal restart strategy. The cheaper the relative cost of going down
the search-tree, the cheaper it is to perform a restart. Therefore, it is expected that future improvements in
BCP speed and additional reasoning will ensure that the optimal restart strategy will be more rapid.

\subsection{Complete local search}
\label{sec:cls}

Ten challenges for SAT solving have been posed by Selman et al.~\cite{Ten} in 1997. Although several
of these challenges have been faced, hardly any progress has been reported on Challenge 5: Desiging
a competitive complete local search solver. Appearently, it is hard to add completeness to local
search solvers effectively. On the other hand, CDCL solvers have been slowly begun to mimic local search 
solvers. This could explain why the performance of current CDCL solvers heavily depends on the seed,
even for unsatisfiable benchmarks.

An important step towards local search is the introduction of phase-saving in CDCL solvers in 
2007~\cite{Pipatsrisawat2007}. Essentially the same technique is used in the local search SAT solver
{\tt UnitWalk}~\cite{UnitWalk} since 2001. The UnitWalk algorithm
starts by initializing a random full assignment. In each iteration, this assignment is improved by the 
following procedure: First, a random order of the variables is created. Second, the most important 
(based on this random order) free variable is assigned to the value in the full assignment. Third,
BCP is applied. Each assignment due to BCP is copied to the full assignment.
After BCP is finished, the procedure returns to the second step until all variables are assigned. 
In other words, both techniques copy the value {\em from} the full assignment for decision variables 
and copy the value {\em to} the full assignment for implied variables.

Due to the combination of phase-saving and rapid restarts, one can argue that CDCL solvers hardly
perform search anymore. They merely improve the full assignment, while recording clauses for 
every encountered conflict. Therefore, modern CDCL solvers could be considered as complete 
local search solvers. This claim will become stronger if the trend towards ultra rapid restarts will
continue.

\section{Results}

Because we observed several signs in favor of ultra rapid restarts, we decided to experiment with strategies
that restart radically faster compared to those used in the current CDCL solvers.
The dataset we have used for this experiment includes all industrial instances which were used in the 
SAT competition of 2009. All the experiments presented in this paper were conducted on a cluster of 
14 Dual Intel Xeon CPUs with EM64T (64-bit) extensions, running at 3.2GHz with 4GB of RAM under 
Debian GNU/Linux 4.0. 

The solver we used for the experiments is the award-winning {\tt MiniSAT} 2.0 which we equipped with a 
phase-saving direction heuristic.
We experimented with 12 different unit runs for the Luby sequence and used a timeout of 900 seconds. To provide 
more stable numbers, we ran all experiments with three different seeds.

\begin{figure}[h]
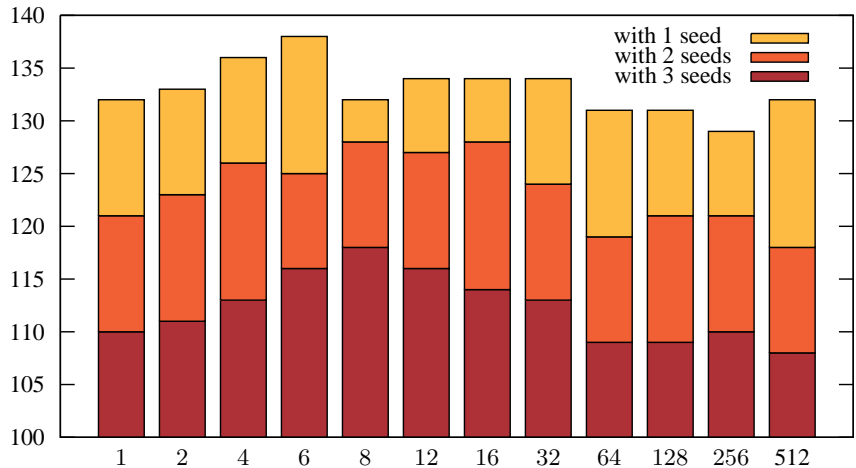

\vspace{-5pt}
\include{"ps_luby"}
\vspace{-15pt}
\caption{Histrogram showing the number of solved instances for different unit runs of the Luby sequence. 
The baseline at 100 instances represents the original version of {\tt MiniSAT} 2.0.}
\label{fig:luby}
\end{figure}

\newpage

The original version of {\tt MiniSAT} 2.0, which applies negative branching and uses a geometrical restart strategy, 
solves 100 instances (44 SAT, 56 UNSAT) within the timeout. Figure~\ref{fig:luby} shows the results for the 
adapted solver using phase-saving and Luby sequences. Notice that using any of these Luby sequences solves many
more benchmark instances compared to the original version. The optimal restart strategy for this test set seems
around a unit run of 6 or 8. Recall that this number is much smaller than what is commonly used in CDCL solvers.

The size of the unit run has a clear impact on the number of conflicts the solver encounters while solving 
a problem. Table~\ref{Table:conflicts} shows the average numbers. The smaller the unit run, the smaller the
number of conflicts. Although the results using a unit run of 1 and 512 show a comparable performance on
the dataset, the former resolves significantly fewer conflicts. Apparently, smaller unit runs require less
search to solve instances. The results using other unit runs hint in this direction as well.
Therefore, we expect that -- assuming that the (relative) cost of performing a restart will further be reduced -- 
even smaller unit runs will appear optimal in the future.

\begin{table}
	\centering
	\caption{The average number of conflicts for several unit runs of the Luby sequence.}
	\label{Table:conflicts}

  \begin{tabular}{ @{\hspace{10pt}}l@{\hspace{30pt}} c @{\hspace{20pt}} c @{\hspace{20pt}} c @{\hspace{20pt}} c @{\hspace{20pt}} c @{\hspace{10pt}}}
  \hline 
		
		Strategy & SAT & UNSAT & SOLVED & UNSOLVED & ALL\\
		\hline \hline
		\emph{Luby-1} & 90465	&	171629	&	137513	&	309941	&	237799 \\\hline
		\emph{Luby-2} & 79064	&	181351	&	138151	&	349777	&	260505 \\\hline
		\emph{Luby-4} & 76743	&	188944	&	140772	&	380772	&	277324  \\\hline
		\emph{Luby-6} & 83970	&	204974	&	153252	&	387720	&	285578  \\\hline
		\emph{Luby-8} & 81043	&	210837	&	155211	&	401257	&	294354  \\\hline
		\emph{Luby-12} & 91667	&	197671	&	151839	&	412065	&	299301  \\\hline
		\emph{Luby-16} & 88195	&	205252	&	153884	&	428446	&	309785 \\\hline
		\emph{Luby-32} & 95870	&	222315	&	167783	&	436521	&	321921 \\\hline
		\emph{Luby-64} & 79550	&	212722	&	155225	&	452032	&	329556 \\\hline
		\emph{Luby-128} & 93769	&	214950	&	160234	&	470681	&	341863 \\\hline
		\emph{Luby-256} & 96148	&	222214	&	165134	&	477443	&	348211 \\\hline
		\emph{Luby-512} & 95981	&	222871	&	164034	&	487855	&	354604  \\\hline
		                           		                                              		                           		
  \end{tabular}
\end{table}

\section{Conclusions}

We showed that the award winning solver {\tt MiniSAT} 2.0 can significantly be improved by adding phase-saving
and rapid restarts. The optimal strategy on the industrial benchmarks of the SAT 2009 competition restarts
far more frequently compared to strategies used by the current state-of-the-art solvers. This result supports
our observation that SAT solvers tend towards ultra rapid restarts and become complete local search solvers.

\newpage

\bibliographystyle{splncs}
\bibliography{rapid}

\end{document}

%% file: ps_luby.tex
\ifx\PSTloaded\undefined
\def\PSTloaded{t}
\psset{arrowsize=.01 3.2 1.4 .3}
\psset{dotsize=.01}
\catcode`@=11

\newpsobject{PST@Border}{psline}{linewidth=.0015,linestyle=solid}
\newpsobject{PST@Axes}{psline}{linewidth=.0015,linestyle=dotted,dotsep=.004}
\newpsobject{PST@Solid}{psline}{linewidth=.0015,linestyle=solid,fillstyle=solid,fillcolor=Maroon}
\newpsobject{PST@Dashed}{psline}{linewidth=.0015,linestyle=solid,fillstyle=solid,fillcolor=RedOrange}
\newpsobject{PST@Dotted}{psline}{linewidth=.0015,linestyle=solid,fillstyle=solid,fillcolor=Dandelion}
\newpsobject{PST@LongDash}{psline}{linewidth=.0015,linestyle=dashed,dash=.02 .01}
\newpsobject{PST@Diamond}{psdots}{linewidth=.001,linestyle=solid,dotstyle=square,dotangle=45}
\newpsobject{PST@Filldiamond}{psdots}{linewidth=.001,linestyle=solid,dotstyle=square*,dotangle=45}
\newpsobject{PST@Cross}{psdots}{linewidth=.001,linestyle=solid,dotstyle=+,dotangle=45}
\newpsobject{PST@Plus}{psdots}{linewidth=.001,linestyle=solid,dotstyle=+}
\newpsobject{PST@Square}{psdots}{linewidth=.001,linestyle=solid,dotstyle=square}
\newpsobject{PST@Circle}{psdots}{linewidth=.001,linestyle=solid,dotstyle=o}
\newpsobject{PST@Triangle}{psdots}{linewidth=.001,linestyle=solid,dotstyle=triangle}
\newpsobject{PST@Pentagon}{psdots}{linewidth=.001,linestyle=solid,dotstyle=pentagon}
\newpsobject{PST@Fillsquare}{psdots}{linewidth=.001,linestyle=solid,dotstyle=square*}
\newpsobject{PST@Fillcircle}{psdots}{linewidth=.001,linestyle=solid,dotstyle=*}
\newpsobject{PST@Filltriangle}{psdots}{linewidth=.001,linestyle=solid,dotstyle=triangle*}
\newpsobject{PST@Fillpentagon}{psdots}{linewidth=.001,linestyle=solid,dotstyle=pentagon*}
\newpsobject{PST@Arrow}{psline}{linewidth=.001,linestyle=solid}
\catcode`@=12

\fi
\psset{unit=5.0in,xunit=5.0in,yunit=2.5in}
\pspicture(0.000000,0.000000)(1.000000,1.000000)
\ifx\nofigs\undefined
\catcode`@=11

\PST@Border(0.1170,0.0840)
(0.1320,0.0840)

\PST@Border(0.9470,0.0840)
(0.9320,0.0840)

\rput[r](0.1010,0.0840){ 100}
\PST@Border(0.1170,0.1945)
(0.1320,0.1945)

\PST@Border(0.9470,0.1945)
(0.9320,0.1945)

\rput[r](0.1010,0.1945){ 105}
\PST@Border(0.1170,0.3050)
(0.1320,0.3050)

\PST@Border(0.9470,0.3050)
(0.9320,0.3050)

\rput[r](0.1010,0.3050){ 110}
\PST@Border(0.1170,0.4155)
(0.1320,0.4155)

\PST@Border(0.9470,0.4155)
(0.9320,0.4155)

\rput[r](0.1010,0.4155){ 115}
\PST@Border(0.1170,0.5260)
(0.1320,0.5260)

\PST@Border(0.9470,0.5260)
(0.9320,0.5260)

\rput[r](0.1010,0.5260){ 120}
\PST@Border(0.1170,0.6365)
(0.1320,0.6365)

\PST@Border(0.9470,0.6365)
(0.9320,0.6365)

\rput[r](0.1010,0.6365){ 125}
\PST@Border(0.1170,0.7470)
(0.1320,0.7470)

\PST@Border(0.9470,0.7470)
(0.9320,0.7470)

\rput[r](0.1010,0.7470){ 130}
\PST@Border(0.1170,0.8575)
(0.1320,0.8575)

\PST@Border(0.9470,0.8575)
(0.9320,0.8575)

\rput[r](0.1010,0.8575){ 135}
\PST@Border(0.1170,0.9680)
(0.1320,0.9680)

\PST@Border(0.9470,0.9680)
(0.9320,0.9680)

\rput[r](0.1010,0.9680){ 140}
\PST@Border(0.1808,0.0840)
(0.1808,0.1040)


\rput(0.1808,0.0420){$1$}
\PST@Border(0.2447,0.0840)
(0.2447,0.1040)


\rput(0.2447,0.0420){$2$}
\PST@Border(0.3085,0.0840)
(0.3085,0.1040)


\rput(0.3085,0.0420){$4$}
\PST@Border(0.3724,0.0840)
(0.3724,0.1040)


\rput(0.3724,0.0420){$6$}
\PST@Border(0.4362,0.0840)
(0.4362,0.1040)


\rput(0.4362,0.0420){$8$}
\PST@Border(0.5001,0.0840)
(0.5001,0.1040)


\rput(0.5001,0.0420){$12$}
\PST@Border(0.5639,0.0840)
(0.5639,0.1040)


\rput(0.5639,0.0420){$16$}
\PST@Border(0.6278,0.0840)
(0.6278,0.1040)


\rput(0.6278,0.0420){$32$}
\PST@Border(0.6916,0.0840)
(0.6916,0.1040)


\rput(0.6916,0.0420){$64$}
\PST@Border(0.7555,0.0840)
(0.7555,0.1040)


\rput(0.7555,0.0420){$128$}
\PST@Border(0.8193,0.0840)
(0.8193,0.1040)


\rput(0.8193,0.0420){$256$}
\PST@Border(0.8832,0.0840)
(0.8832,0.1040)


\rput(0.8832,0.0420){512}
\PST@Border(0.1170,0.9680)
(0.1170,0.0840)
(0.9470,0.0840)
(0.9470,0.9680)
(0.1170,0.9680)

\rput[r](0.8200,0.9270){with 1 seed\,~}
\PST@Solid(0.84,0.85)(0.84,0.83)
(0.90,0.83)(0.90,0.85)(0.84,0.85)
\PST@Solid(0.1569,0.0840)
(0.1569,0.3050)
(0.2048,0.3050)
(0.2048,0.0840)
(0.1569,0.0840)

\PST@Solid(0.2207,0.0840)
(0.2207,0.3271)
(0.2686,0.3271)
(0.2686,0.0840)
(0.2207,0.0840)

\PST@Solid(0.2846,0.0840)
(0.2846,0.3713)
(0.3325,0.3713)
(0.3325,0.0840)
(0.2846,0.0840)

\PST@Solid(0.3484,0.0840)
(0.3484,0.4376)
(0.3963,0.4376)
(0.3963,0.0840)
(0.3484,0.0840)

\PST@Solid(0.4123,0.0840)
(0.4123,0.4818)
(0.4602,0.4818)
(0.4602,0.0840)
(0.4123,0.0840)

\PST@Solid(0.4761,0.0840)
(0.4761,0.4376)
(0.5240,0.4376)
(0.5240,0.0840)
(0.4761,0.0840)

\PST@Solid(0.5400,0.0840)
(0.5400,0.3934)
(0.5879,0.3934)
(0.5879,0.0840)
(0.5400,0.0840)

\PST@Solid(0.6038,0.0840)
(0.6038,0.3713)
(0.6517,0.3713)
(0.6517,0.0840)
(0.6038,0.0840)

\PST@Solid(0.6677,0.0840)
(0.6677,0.2829)
(0.7156,0.2829)
(0.7156,0.0840)
(0.6677,0.0840)

\PST@Solid(0.7315,0.0840)
(0.7315,0.2829)
(0.7794,0.2829)
(0.7794,0.0840)
(0.7315,0.0840)

\PST@Solid(0.7954,0.0840)
(0.7954,0.3050)
(0.8432,0.3050)
(0.8432,0.0840)
(0.7954,0.0840)

\PST@Solid(0.8592,0.0840)
(0.8592,0.2608)
(0.9071,0.2608)
(0.9071,0.0840)
(0.8592,0.0840)

\rput[r](0.8200,0.8850){with 2 seeds}
\PST@Dashed(0.84,0.89)(0.84,0.87)
(0.90,0.87)(0.90,0.89)(0.84,0.89)
\PST@Dashed(0.1569,0.3050)
(0.1569,0.5481)
(0.2048,0.5481)
(0.2048,0.3050)
(0.1569,0.3050)

\PST@Dashed(0.2207,0.3271)
(0.2207,0.5923)
(0.2686,0.5923)
(0.2686,0.3271)
(0.2207,0.3271)

\PST@Dashed(0.2846,0.3713)
(0.2846,0.6586)
(0.3325,0.6586)
(0.3325,0.3713)
(0.2846,0.3713)

\PST@Dashed(0.3484,0.4376)
(0.3484,0.6365)
(0.3963,0.6365)
(0.3963,0.4376)
(0.3484,0.4376)

\PST@Dashed(0.4123,0.4818)
(0.4123,0.7028)
(0.4602,0.7028)
(0.4602,0.4818)
(0.4123,0.4818)

\PST@Dashed(0.4761,0.4376)
(0.4761,0.6807)
(0.5240,0.6807)
(0.5240,0.4376)
(0.4761,0.4376)

\PST@Dashed(0.5400,0.3934)
(0.5400,0.7028)
(0.5879,0.7028)
(0.5879,0.3934)
(0.5400,0.3934)

\PST@Dashed(0.6038,0.3713)
(0.6038,0.6144)
(0.6517,0.6144)
(0.6517,0.3713)
(0.6038,0.3713)

\PST@Dashed(0.6677,0.2829)
(0.6677,0.5039)
(0.7156,0.5039)
(0.7156,0.2829)
(0.6677,0.2829)

\PST@Dashed(0.7315,0.2829)
(0.7315,0.5481)
(0.7794,0.5481)
(0.7794,0.2829)
(0.7315,0.2829)

\PST@Dashed(0.7954,0.3050)
(0.7954,0.5481)
(0.8432,0.5481)
(0.8432,0.3050)
(0.7954,0.3050)

\PST@Dashed(0.8592,0.2608)
(0.8592,0.4818)
(0.9071,0.4818)
(0.9071,0.2608)
(0.8592,0.2608)

\rput[r](0.8200,0.8430){with 3 seeds}
\PST@Dotted(0.84,0.93)(0.84,0.91)
(0.90,0.91)(0.90,0.93)(0.84,0.93)
\PST@Dotted(0.1569,0.5481)
(0.1569,0.7912)
(0.2048,0.7912)
(0.2048,0.5481)
(0.1569,0.5481)

\PST@Dotted(0.2207,0.5923)
(0.2207,0.8133)
(0.2686,0.8133)
(0.2686,0.5923)
(0.2207,0.5923)

\PST@Dotted(0.2846,0.6586)
(0.2846,0.8796)
(0.3325,0.8796)
(0.3325,0.6586)
(0.2846,0.6586)

\PST@Dotted(0.3484,0.6365)
(0.3484,0.9238)
(0.3963,0.9238)
(0.3963,0.6365)
(0.3484,0.6365)

\PST@Dotted(0.4123,0.7028)
(0.4123,0.7912)
(0.4602,0.7912)
(0.4602,0.7028)
(0.4123,0.7028)

\PST@Dotted(0.4761,0.6807)
(0.4761,0.8354)
(0.5240,0.8354)
(0.5240,0.6807)
(0.4761,0.6807)

\PST@Dotted(0.5400,0.7028)
(0.5400,0.8354)
(0.5879,0.8354)
(0.5879,0.7028)
(0.5400,0.7028)

\PST@Dotted(0.6038,0.6144)
(0.6038,0.8354)
(0.6517,0.8354)
(0.6517,0.6144)
(0.6038,0.6144)

\PST@Dotted(0.6677,0.5039)
(0.6677,0.7691)
(0.7156,0.7691)
(0.7156,0.5039)
(0.6677,0.5039)

\PST@Dotted(0.7315,0.5481)
(0.7315,0.7691)
(0.7794,0.7691)
(0.7794,0.5481)
(0.7315,0.5481)

\PST@Dotted(0.7954,0.5481)
(0.7954,0.7249)
(0.8432,0.7249)
(0.8432,0.5481)
(0.7954,0.5481)

\PST@Dotted(0.8592,0.4818)
(0.8592,0.7912)
(0.9071,0.7912)
(0.9071,0.4818)
(0.8592,0.4818)

\PST@Border(0.1170,0.9680)
(0.1170,0.0840)
(0.9470,0.0840)
(0.9470,0.9680)
(0.1170,0.9680)

\catcode`@=12
\fi
\endpspicture